\documentclass[sigconf]{acmart}
\usepackage{graphics}
\usepackage{multirow}
\usepackage{enumitem}
\setlist[description]{style=nextline}
\AtBeginDocument{%
  \providecommand\BibTeX{{%
    \normalfont B\kern-0.5em{\scshape i\kern-0.25em b}\kern-0.8em\TeX}}}

\acmConference[KDD Workshop on Machine Learning in Finance, 2023]{KDD Workshop on Machine Learning in Finance, 2023}{August 06-10, 2023}{Long Beach, CA}

\setcopyright{acmcopyright}
\copyrightyear{2023}
\acmYear{2023}
\acmDOI{XXXXXXX.XXXXXXX}

\acmPrice{15.00}
\acmISBN{978-1-4503-XXXX-X/18/06}

\begin{document}

\title{An Ensemble Approach to Personalized Real Time Predictive Writing for Experts}



 \author{Sourav Prosad}
 \email{sourav_prosad@intuit.com}
 \affiliation{%
  \institution{Intuit AI, Bengaluru}
  \country{India}
 }

 \author{Viswa Datha Polavarapu}
 \email{vishwadatha_p@intuit.com}
 \affiliation{%
  \institution{Intuit AI, Bengaluru}
  \country{India}
 }

 \author{Shrutendra Harsola}
 \email{shrutendra_harsola@intuit.com}
 \affiliation{%
  \institution{Intuit AI, Bengaluru}
  \country{India}
 }







\renewcommand{\shortauthors}{S. Prosad, V. Datta and S. Harsola}


\begin{abstract}
  Completing a sentence, phrase or word after typing few words / characters is very helpful for Intuit financial experts, while taking notes or having a live chat with users, since they need to write complex financial concepts more efficiently and accurately many times in a day. 
  In this paper, we tie together different approaches like large language models, traditional Markov Models and char level models to create an end-to-end system to provide personalised sentence/word auto-complete suggestions to experts, under strict latency constraints. Proposed system can auto-complete sentences, phrases or words while writing with personalisation and can be trained with very less data and resources with good efficiency. Our proposed system is not only efficient and personalized but also robust as it leverages multiple machine learning techniques along with transfer learning approach to fine tune large language model with Intuit specific data. This ensures that even in cases of rare or unusual phrases, the system can provide relevant auto-complete suggestions in near real time. 
  Survey has showed that this system saves expert note-taking time and boosts expert confidence in their communication with teammates and clients.
  Since enabling this predictive writing feature for QBLive experts, more than a million keystrokes have been saved based on these suggestions. We have done comparative study for our ensemble choice. Moreover this feature can be integrated with any product which has writing facility within a very short period of time.
\end{abstract}

\ccsdesc[500]{Computing methodologies~Natural language generation}



\keywords{Language Model, Assisted Writing, Sentence Auto-Completion}


\maketitle

\section{Introduction}
\textbf{Intuit Experts: } Intuit experts help customers with book keeping and tax preparation and other complex financial tasks. These experts use expert portal to keep track of their work schedule and connect with their team members, as well as customers. They also use experts portal to take notes to keep track of their engagement and also, to chat with the customers. So, they need to type a lot of text in a day. But, the context of the notes or chats are within very limited domain of book keeping, tax preparation and financial issues. For that reason, they type very similar words or similar sentences repeatedly in a day. Our goal is to provide text auto-completion in the experts portal to improve the writing experience for the experts.

\textbf{Text auto-completion: } Smart reply \cite{kannan2016smart} and Smart compose \cite{chen2019gmail} are two recent works that aims to solve the similar problem for emails. But their proposed approach has couple of challenges when applied to Intuit experts domain. First, they train the models from scratch requiring huge amount of data, time and computational resources, which is difficult in different organisations and enable this feature in their product in a very short span. Second, they focus only on suggesting next few words based on the words written so far. They do not address the task of also suggesting word completions based on first few chars. Third, they have not experimented with latest neural language models.

\textbf{Neural Language models: } Auto-regressive neural language models like GPT2 ~\cite{radford2019language} are well suited for such text auto-completion tasks. But these models are not well suited for personalized text auto-completion, especially when trained in transfer learning setting. \citet{bakhtin2018lightweight} aims to address the problem of personalization by training a Seq2Seq model from scratch, and concatenating user profile information with the input text. But using a single model to capture individual expert's writing style is not efficient, since it requires a huge amount of data and resources to train the model.

\vspace{1em}
\textbf{Contributions: } Main contributions of the paper are:
\begin{itemize}
    \item Proposed an ensemble of global neural language model using transfer learning, local Markov models to provide text completion suggestions after every word, with personalization and strict latency limits.
    \item Proposed an approach combining word level ensemble model results with char level language model to provide text completion suggestions after every character.
    \item Extensive experiments to compare various modeling choices
    \item Description of production deployment of the end-to-end system.
\end{itemize}

\section{Related Work}
\textbf{Language Models:} Real time predictive writing is an usecase of language modelling task. It helps user to complete phrases and improves their writing efficiency. Language modelling is extensively used for downstream tasks of NLP. There are many different techniques available for language modelling. Sequence to sequence model is very widely used method for machine translation(~\cite{sutskever2014sequence}, ~\cite{firat2016multi},~\cite{johnson2017google}, ~\cite{lample2019cross}). In recent days Neural language modelling (~\cite{dauphin2017language}, ~\cite{bengio2000neural}, ~\cite{jozefowicz2016exploring},~\cite{melis2017state},
~\cite{mikolov2010recurrent}, ~\cite{matthew2018peters}, 
~\cite{yang2017breaking}, 
~\cite{zolna2017fraternal}) is very popular which improves the performance of state of the art techniques. n-gram langauge modelling (~\cite{kneser1995improved},~\cite{vidal1996using}) is also heavily used technique to solve NLP problems. The performance of n-gram models are inferior than state of the art neural language modelling when the dataset is huge. But it is very useful technique for small data. 

\textbf{Transformer Architecture:} In recent times, transformer architecture becomes very popular for NLP tasks. Transformers are pre-trained with huge amount of data for general purpose use. GPT2(~\cite{radford2019language}) is decoder based transformer which is pre-trained on a very large corpus of English data in a self-supervised fashion. It outperforms many neural language modelling techniques of language modelling. 

\textbf{Predictive Writing in Industry:} Most related to Personalized Real Time Predictive Writing is Smart-Compose(\cite{chen2019gmail}), which takes current written prefix and older emails as input to the model and tries to generate next sequence of words. Personalized Real Time Predictive Writing is different from Smart-compose in several ways. Firstly, Smart compose uses LSTM based encoder-decoder model to generate the suggestions, whereas Personalized Real Time Predictive Writing leverages the power of self-attention from transformer architecture and fine-tuned GPT2 with specific data to generate output in a very short time which can be used in industry with less resources. Secondly, a RNN based char level language model is used for word completion in Personalized Real Time Predictive Writing unlike smart-compose. RNN based character level language model is very easy to train with very less data.

\textbf{Personalization:} Personalising the suggestion for each user is considered to be a very important feature for completing sentences(~\cite{hsu2007generalized}, ~\cite{jaech2018personalized}, ~\cite{lee2016personalizing}, ~\cite{tseng2015personalizing}, ~\cite{yoon2017efficient}). Personalized Real Time Predictive Writing uses ensemble techniques of transformer based language model and k-order Markov Model to predict the next few words given a input prefix which is discussed in ~\cite{bakhtin2018lightweight}. Interpolation between k-order Markov Model and neural network probabilities of every word is being used for personalising responses.

\section{Text Completion after every word}
\label{sec:method}

\subsection{Model Architecture}
Personalized Real Time Predictive Writing is a language modelling task. One of the most important features of the model is to predict the phrases/words based on different users and their writing style. 
We formulated the ML problem as language modelling problem, where, for a user j, given a sequence \[S = \{ s_1, s_2, s_3,...s_n \}\] over the vocabulary V, we seek to model
\[p(S|user_j) = \prod_{t=1}^n p(s_t|s_{<t};user_j)\]
where $s_{<t}$ denotes $s_1, s_2,...s_{t-1}$

Next, we decompose this probability into following two terms:

\[= \alpha* \prod_{t=1}^n p(s_t|s_{<t}) +(1-\alpha)* \prod_{t=k+1}^n p(s_{t}|s_{t-1:t-k}:user_j)\]

where, first term is independent of the user and will be estimated by a global language model trained on all users data. Second term aims to incorporate the personalized writing style of each user and will be estimated by training separate lightweight local language models for each user. Finally, $\alpha \epsilon [0,1] $ is the weight we want to put on our global language model and can be chosen as described in ~\cite{bakhtin2018lightweight}.

Overall text auto-completion system after every word, consists of the following:
\begin{itemize}
\item {\bfseries Global Language Model}: An auto-regressive large neural language model (GPT2), fine-tuned on domain specific data is used as Global model.
\item {\bfseries Local and Personalized kth order Markov models}: A personalized k-th order Markov Model (ngram model) is trained on the go with the data of every expert to capture the writing pattern of that particular expert.
\end{itemize}

\subsection{Data Generation}
Experts historical data is used to train our model. Following pre-processing steps are followed before passing data to the model:
\begin{itemize}
\item {\bfseries decontract}: Decontracted the acronyms i.e. ain't = is not
\item {\bfseries remove dates}: removed dates, months, years etc and replaced it with special token <date>
\item {\bfseries normalize text}: text has been normalized by removing sensitive customer and financial information, urls, tags, credit card information, html tags and replaced these with different special tokens like <user>, <url> etc.
\end{itemize}

We tokenized the text using byte-level Byte-Pair-Encoding. We used starting and ending token to capture starting and ending of a sentence. Padding is also done. Then all the notes are concatenated together. Then we split them in examples of a certain sequence length or block-size. This way the model will receive chunks of contiguous text that may look like:
\verb|part of text 1|
or
\verb|end of text 1<endoftext> <startoftext> beginning of text 2|
depending on whether they span over several of the original texts in the dataset or not. The labels will be the same as the inputs, shifted to the left.
We used training token size of 21446759.

\subsection{Global Language Model}
\label{sub:global}
For global language model, we used auto-regressive neural language model, since they have shown great success for global text generation. Specifically, we used GPT2 architecture in our system, since it provides required trade-off between scale and performance (P99 inference latency around 100ms, more details in experiments section). Having said that, our proposed system is modular in nature and GPT2 can easily be swapped with any other global language model, if it shows better results under given latency constraints.

We started with pre-trained GPT2 ~\cite{radford2019language} tokenizer and model, and then, fine-tuned the model on our financial data using transfer learning. This fine-tuning process allows the model to learn the specific language and patterns of financial data, improving its accuracy and effectiveness for text completion tasks. Fig. ~\ref{fig:gpt2_architecture} shows the architecture of GPT2. For fine-tuning, we split the data into 80\% train and 20\% validation data split. Training data was first tokenized using pre-trained tokenizer and concatenated together, then split in small chunks with block-size of 256 tokens. Model input is the chunk and the labels will be the same as the inputs, shifted to the left. Next, this inputs and labels are fed into the model and training was parallelized on GPU instances.

Training parameters are as follows:

\begin{figure}[h]
  \includegraphics[width=0.45\textwidth]{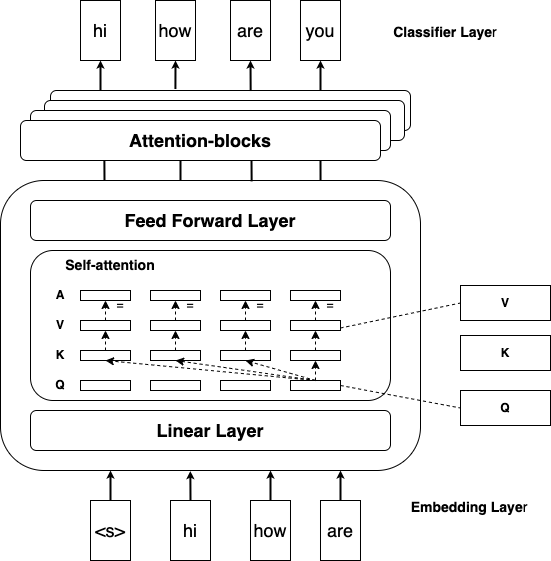}
  \caption{GPT2 Architecture}
  \label{fig:gpt2_architecture}
\end{figure}

\begin{itemize}
\item {\bfseries block\_size}: 256
\item {\bfseries num\_train\_epochs}: 6
\item {\bfseries learning\_rate}: 0.001
\item {\bfseries weight\_decay}: 0.01
\item {\bfseries warmup\_ratio}: 0.1
\item {\bfseries lr\_scheduler\_type}: linear
\item {\bfseries GPU instance}: AWS ml.p3.8xlarge(4 GPU, 32 vCPUs, 244 GiB of memory, 10 Gbps network performance)
\end{itemize}

training loss is 1.34 and validation loss is 1.74 with perplexity 5.71.

\subsection{Local and Personalized Language Models}
\label{sub:local}
Given a sequence of last k words, a k-order Markov model predicts the most probable word that might follow this sequence.
For n>0, \[p(S) = \prod_{t=k+1}^n p(s_{t}|s_{t-1:t-k}:user_j)\]

A k-order Markov language model is also trained on the go for every user to capture the writing style of a particular expert. It will be a local model for that user. 

The way k-order Markov Model computes probability is very intuitive. We compute the maximum likelihood estimation for the parameters of a Markov Model by getting the counts from a corpus and normalize those counts to get probabilities between 0 and 1.

Lets say, we want to compute a tri-gram probability of a word $S_n$ given its previous words $S_{n-1}$ and $S_{n-2}$. The tri-gram count C($S_{n-2}$ $S_{n-1}$ $S_n$) will be normalized by bi-gram count C($S_{n-2}$ $S_{n-1}$). The likelihood function is as following:
\[P(S_n|S_{n-1} S_{n-2})=C(S_{n-2} S_{n-1} S_n)/C(S_{n-2} S_{n-1})\]

\subsection{Word level Ensemble Model}
\label{sub:ensemble}
From the global model we will get a list of suggestions with assigned probabilities and from the local model also we will get a list of suggestions with assigned probabilities. We then combine these two lists with a particular weight to get the final probabilities of each word in the vocabulary.
\[p(s_t|s_{<t};user_j) = \alpha* p_{global}(s_t|s_{<t}) + (1-\alpha)*  p_{local}(s_{t}|s_{t-1:t-k};user_j)\]

Probability will be normalized with a custom normalizer based on the length of the predicted suggestions. 
\[normalizer = (((5 + len\_seq) ** \alpha)/((5 + 1) ** \alpha)) \]
\[ log\_probability = log\_probability/normalizer \]

where len\_seq is the total length of output sequence and $\alpha$ is a constant. In our experiment $\alpha = 0.4$. $\alpha$ is chosen by grid-search to maximize the ExactMatchRate  [~\ref{sub:evaluation_metrics}].

The word with the highest probability is chosen for suggestion. If that crosses a predefined threshold , we will add this word to the final prediction.

In final production 2-order Markov Model is used and last 3 months data of every user is taken to train the personalized model.

\begin{figure}[h]
  \includegraphics[width=0.45\textwidth]{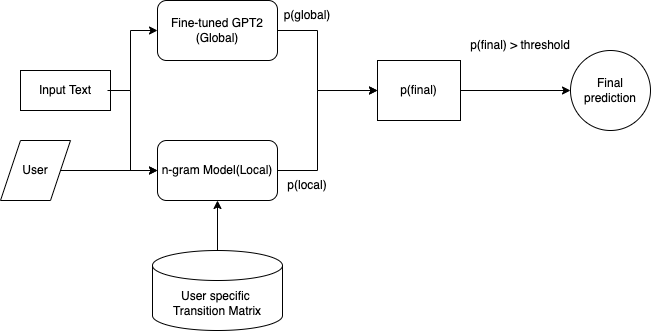}
  \caption{Combination of global and local model}
  \label{fig:ensemble}
\end{figure}

Fig ~\ref{fig:ensemble} shows how the global and the local model are combined. k-order Markov Model is taking user information and user specific transition matrix along with the input text to generate text. We are storing user metadata in user specific transition matrix.

\section{Text Completion after every character}
In previous section, we discussed how a combination of global and local models is used to provide text completion suggestion after every word. In this section, we will talk about how we can further enhance the system to provide text completion suggestions after every character that user types.

\subsection{Caching suggestions from word level ensemble model}
Performing model inference after every character can be very costly, especially in a real time latency sensitive task like text auto-completion. To handle such scenario, we cache suggestions from word level ensemble model at the UI client side. Essentially, word level ensemble model is invoked after every word and instead of returning only top most suggestion, word level model returns a list of top n suggestions to the UI client. Top most suggestion is immediately presented to the user, while the rest of the suggestions are cached at the client side. If the user doesn't accept the top most suggestion and starts typing, then we try to do a prefix match between the characters typed by the user and list of cached suggestions. If a match is found, then the corresponding suggestion is immediately presented to the user. This approach allows us to provide very interactive text completion experience to the users, without increasing the cost significantly.

\subsection{Char Level Language Model}

\begin{figure}[h]
  \includegraphics[width=0.45\textwidth]{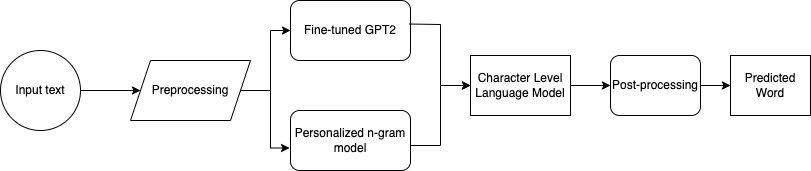}
  \caption{Inference Workflow of Auto-Completion}
  \label{fig:architecture}
\end{figure}

To further increase the coverage of suggestion provided after every character, we have developed a LSTM based character level language model. Essentially, if the characters typed by the user does not match with any of the cached suggestions, then this char level model is invoked. This char level neural language model is trained on historical Experts data and has a P99 latency of under 40ms.

\begin{figure}[h]
  \includegraphics[width=0.45\textwidth]{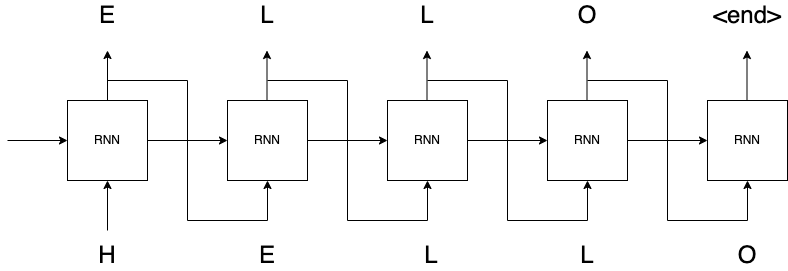}
  \caption{Char level RNN}
  \label{fig:word completion architecture}
\end{figure}

\section{Production Deployment}
\begin{figure}[h]
  \includegraphics[width=0.45\textwidth]{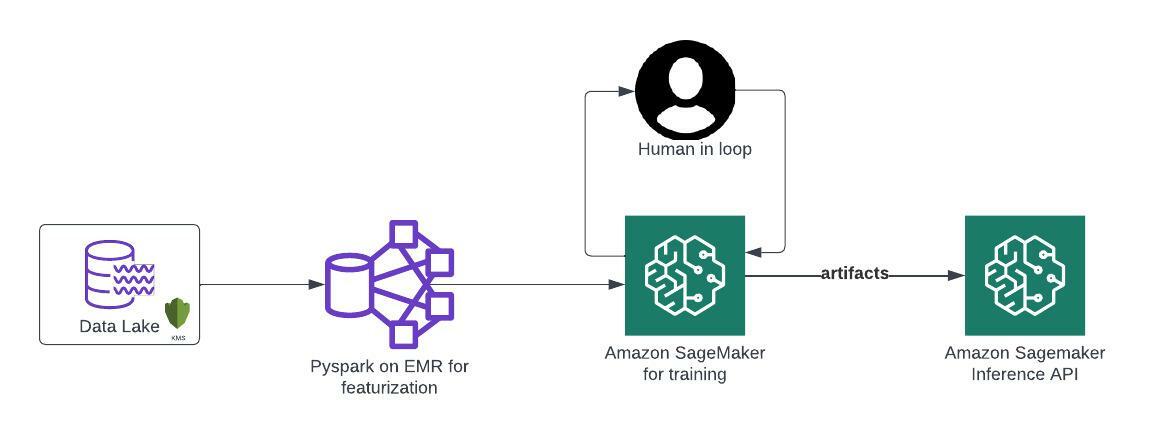}
  \caption{Model deployment architecture}
  \label{fig:Model deployment architecture}
\end{figure}

We have set up the data privacy, data processing, model training and model inference as an end-to-end automated pipeline. Figure \ref{fig:Model deployment architecture} illustrates the model deployment architecture and below, we detail each of the steps involved. 

\textbf{Featurization:} 
Features for about ~15 million past memos of customer care experts from Intuit datalake are generated and used for training of our model. The encrypted features are stored in AWS S3 and we use PySpark for feature processing due to its proven performance characteristics at scale as well as familiarity with scientists and engineers alike. The batch nature of job allows for massive parallelization across executors with ability to tune for cost and speed. 

\textbf{Model training pipeline:} The training is done in PyTorch framework on a GPU machine. All the training and inference code is packaged into their own containers with a continuous integration pipeline. The model is automated to retrain on schedule with a human in loop to review the model metrics before replacing a new model in production.

\textbf{Real-time inference:} Real-time inference is done using AWS SageMaker Inference with GPU instance. Since we are dealing with an auto complete model for customer care agents who inherently type fast, optimized latency is a must. By using greedy approach we achieved a latency of ~80ms with a minimal loss in accuracy. To optimize for user experience, the model's suggestions are provided after an expert completes a word for sentence completion and after two letters of a word are typed for word completion. Figure ~\ref{fig:ensemble} depicts the inference flow given an input text.

\textbf{Feedback collection:} We built a REST API to collect the feedback in real time for every acceptance or rejection from the user. This feedback is stored in the datalake and is used to calculate model performance and make any improvements to the model as required. 

\textbf{Data Privacy:}
We ensure data privacy by encrypting the whole data, both in rest and transit using AWS key management service(KMS) \cite{almeida2019machine}. KMS uses symmetric cryptography and acts as a central repository for storing and governing the keys. It issues the decryption keys to those who have sufficient permissions to do so. During the model training, the data is decrypted on the fly and the generated artifacts are used in model inference. During inference, the collected feedback data is encrypted using the KMS mechanism described above.

\section{EXPERIMENTATION RESULTS}

\subsection{Evaluation Metrics}
\label{sub:evaluation_metrics}
We compare various models against a list of user centric evaluation metrics:
\begin{itemize}

\item {\bfseries Exact Match Rate}: Given an input prefix, model provide text suggestions (word/phrase/sentence) based on the probability threshold discussed in ~\ref{sub:ensemble}. Evaluation has been done to check in how many cases the suggested text is exactly same with the ground truth text (i.e. what was actually written by the user). 

Exact Match Rate = No of Suggestions exactly match with the ground truth text / No of Total Suggestions

\item {\bfseries Effort Saved}: Effort saved is defined as the ratio of character length of accepted suggestions to total character length of the full text. It is an user-assist metric. It defines the percentage of characters written by model instead of expert, which eventually saves the manual effort of typing.

Effort Saved = char length of accepted suggestions / total char length of the full text

\item {\bfseries Coverage}: Coverage is the metric which measures how often suggestion from the model is above threshold. It is measured with respect to the total text length. 

Coverage = No of model calls with suggestion above pre-defined threshold / char length of the full text

\item {\bfseries Avg length of suggestion}: It is defined as the average char length of the suggestions by the model.

Avg length of suggestion = char length of suggestions / No of suggestions
\end{itemize}

\subsection{Experiment setup}
\label{sec: experimental setup}
In this paper, we predominantly use Exact Match Rate and Effort Saved for our evaluation.
The experiment has been done in an experimental setup where we kept different threshold for different combinations of models to ensure fixed coverage from all and computed Exact Match Rate and Effort Saved by predicting next 3 words after giving a user-written prefix as input from randomly selected 50 notes. Fixing the coverage ensures the number of predictions from different models are same. The training and evaluation data-sets are fixed for all combinations of models. Training data has 21.5M tokens and validation data has 5.3M tokens.

\subsection{Text Completion after every word}
\label{sec:exp_word}

\textbf{Choosing the Global model:}
Personalised predictive writing needs to be near real time so that the suggestions pops up very quickly to the users. As described in \cite{nielsen1994usability}, 100 ms is the limit which we can consider as real time. 

Table ~\ref{table:2} compares the latency of different potential global language models. All the models are tested on AWS ml.g4dn.x large instance.
\begin{itemize}
    \item T5-small \cite{t5} , XLNet \cite{xlnet} and BART \cite{bart} have very high inference latency to be considered.
    \item GPT2 has 102 ms $99^{th}$ percentile latency, which is close to the 100 ms constraint that we have.
    \item Global 2nd order Markov model has $99^{th}$ percentile latency of 85 ms.
\end{itemize}

\begin{table}[H]
\begin{tabular}{|p{3.5cm}||p{3.5cm}|}
 \hline
 \textbf{Model} & \textbf{P99 Inference Latency}\\
 \hline
 T5-small &2127 ms\\
 \hline
 XLNet-base-cased &738 ms \\
 \hline
 BART  & 767 ms \\
 \hline
 GPT2  & 102 ms \\
 \hline
 2nd order Markov Model (ngram model)   &85 ms \\
 \hline
\end{tabular}
\caption{Inference Latency Comparison}
\label{table:2}
\end{table}
Considering all the above parameters, next we experimented with 2nd order Markov model and GPT2 as global model for this use case. But our proposed system is modular in nature and base model can be easily replaced with other language model, if it shows better results under given latency constraints. 

\bigskip
\noindent \textbf{Global Language Model Comparison:} Table ~\ref{table:1} compares various modelling options for global language model, in the fixed coverage setup. First, a 2nd-order Markov model (ngram model) is tried, which gives an Exact Match Rate of 21.97 and Effort saved of 5.96. Second, using a pre-trained GPT2 model improves the Exact Match Rate to 29.06 and Effort Saved to 7.55. 

Next we compared with
Smart compose \cite{chen2019gmail}, which is the latest published industrial system for this problem. We experimented with two smart compose architectures, which they concluded as most production-appropriate under given latency constraints. We have used the same training configuration as mentioned in the original paper.
\begin{itemize}
    \item \textbf{Smart-compose LM-A LSTM-2-2048 :} LSTM architecture with 2 layers and 2048 hidden units. It achieved an Exact match rate of 6.38 and effort saved of 2.22.
    \item \textbf{Smart-compose LM-A Transformer-768-2048 :} Transformer architecture with 768 model dimension and 2048 feed-forward inner layer dimension. Model has 6 layers and 8 attention-heads, as proposed in the smart compose paper. It achieved an Exact match rate of 18.27 and effort saved of 4.91.
\end{itemize}
Finally, GPT2 model fine-tuned on our Experts data performs best with Exact match rate of 37.84 and Effort saved of 12.51.

\begin{table}[H]
\begin{tabular}{|p{2.5cm}||p{1cm}|p{1cm}|p{1cm}|p{1cm}|}
 \hline
 Model & Exact-Match-Rate &Coverage &Avg suggestion length& Effort-saved\\
 \hline
 2nd-order Markov Model &21.97   &5 &6.4  &5.96\\
 \hline
 Pre-trained GPT2 &29.06 & 5 &  5.39 & 7.55\\
 \hline
 Smart-compose LM-A LSTM-2-2048  & 6.38  &5 &   5.24 & 2.22 \\
 \hline
 Smart-compose LM-A Transformer-768-2048  & 18.27  &5 &   5.79 & 4.91 \\
 \hline
 Fine-tuned GPT2  & 37.84  &5 &   6.8& 12.51 \\
 \hline
\end{tabular}
\caption{Global Language Model: Fine-tuned GPT2 performs best with ExactMatchRate of 37.84 and Effort Saved of 12.51}
\label{table:1}
\end{table}

Here it can be seen that, due to relatively less training data available (21.5M tokens), fine-tuned GPT2 gives the highest ExactMatchrate with significantly more Effort-Saved than the other models which are trained from scratch (like Markov models and Smart-Compose models). 

\bigskip
\noindent \textbf{Personalized Models Comparison:} We compare following approaches to provide personalized text completion suggestion.

\begin{itemize}
    \item \textbf{2nd-order Markov Model:} Separate 2nd-order Markov Model (ngram model) is trained for each user based on its historical data.
    \item \textbf{Personalized GPT2:} This approach tries to introduce personalization in the GPT2 model itself, by changing the input passed to the GPT2 model both during fine-tuning and inference. Instead of only input text to the GPT2 model, userid is also concatenated with the input text, before passing to the model.
    \item \textbf{Proposed Ensemble Approach :} As proposed in section \ref{sub:ensemble}, this approach combined Global GPT2 language model with local 2nd order Markov models, to provide personalized suggestions.
\end{itemize}

As shown in the Table ~\ref{table:personalized}, our proposed Ensemble approach achieves highest Exact Match Rate of 39.1 and Effort saved of 14.01.

\begin{table}[H]
\begin{tabular}{|p{2.5cm}||p{1cm}|p{1cm}|p{1cm}|p{1cm}|}
\hline
 Model & Exact-Match-Rate &Coverage &Avg suggestion length& Effort-saved\\
 \hline
 2nd-order Markov Model &21.97 &5.05 &6.4 &5.96 \\
 \hline
 Personalized-GPT2  & 35.88 & 5.05 & 7.75 & 13.21\\
 \hline
 Ensemble-Approach &39.1 &5.09 &7.52 &14.01 \\
\hline
\end{tabular}
\caption{Proposed Ensemble Approach achieves highest Exact Match Rate of 39.1 and Effort saved of 14.01.}
\label{table:personalized}
\end{table}

\begin{figure}[h]
  \includegraphics[width=0.45\textwidth]{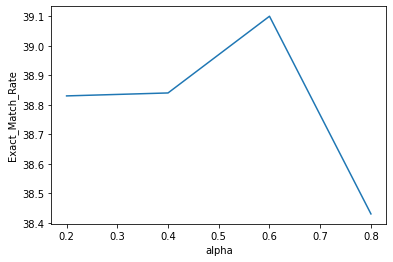}
  \caption{Evaluation of changing alpha with ExactMatchRate}
  \label{fig:Evaluation of changing alpha with ExactMatchrate}
\end{figure}

\bigskip
\noindent \textbf{Selection of Alpha parameter in Ensemble model:}
Alpha is a sensitive parameter when it comes to personalisation. As mentioned in section \ref{sub:ensemble}, alpha ($\alpha$) parameter determines the relative weight between global language model and local personalized models. We did a grid search with alpha values 0.2,0.4,0.6,0.8 to check the performance of the ensemble model. We kept the coverage same while comparing the other metrics to check the effect of alpha. We can see in Fig ~\ref{fig:Evaluation of changing alpha with ExactMatchrate} the performance is at its peak when alpha=0.6 with ExactMatchrate 39.1

Also, with the chosen alpha (0.6), 68.5\% time the results are generated from GPT2 model and 31.4\% time the results are coming from k-th order Markov Model.

\subsection{Text Completion after every character}
\label{sec:exp_char}
In the previous subsection \ref{sec:exp_char}, we experimented with various approaches to provide text completion suggestions after every word and showed that our proposed word level ensemble approach performs best. In this section, we present experiment results for text completion approaches after every character. These approaches are not aimed to replace to text completion approaches presented in previous subsection. Instead they will complement the word level approaches and combined results are presented in next subsection \ref{sec:exp_end_to_end}

Table ~\ref{table:5} shows the results for following approaches:
\begin{itemize}
    \item \textbf{GPT2 with complete text input:} In this approach, input text is passed as it is to the GPT2 model.
    \item \textbf{GPT2 with trimmed text input:} In this approach, input text is trimmed to remove the last incomplete word. Essentially, last few characters are trimmed from the end to reach the last word boundary and the trimmed input is passed to the GPT2. Top 3 suggestions from the GPT2 are then matched with the incomplete word and if a match if found, then that suggestion is presented to the user.
    \item \textbf{Char level Language model:} In this approach, a LSTM based char level language model is trained on historical Experts data.
\end{itemize}

\begin{table}[H]
\begin{tabular}{|p{2.5cm}||p{1cm}|p{1cm}|p{1cm}|p{1cm}|}
\hline
 Model & Exact-Match-Rate &Coverage &Avg suggestion length& Effort-saved\\
 \hline
 GPT2 with complete text input &5.77 &2.88 &3.85 &0.6 \\
 \hline
 GPT2 with trimmed text input &72.57 &2.32 &7.51 &10.3 \\
 \hline
 LSTM based Character level language model &55.29 &2.82 &3.91 &6.57 \\
\hline
\end{tabular}
\caption{Text completion after every character}
\label{table:5}
\end{table}

GPT2 with trimmed text input, performs best for text completion after every character, with an exact match rate of 72.57 and effort saved of 10.53.

\subsection{End-to-end system}
\label{sec:exp_end_to_end}
In this section, we present experiment results for end-to-end system combining approaches for text completion after every word, as well as after every character.

Table ~\ref{table:6} shows that adding text completion after every character improves effort saved by 50\% (14.01 to 21.11), as compared to text completion only after every word.

\begin{table}[H]
\begin{tabular}{|p{1.35cm}|p{1.55cm}||p{0.9cm}|p{0.9cm}|p{0.9cm}|p{0.9cm}|}
\hline
 Text Completion after every word & Text completion after every char & Exact-Match-Rate &Coverage &Avg suggestion length& Effort-saved\\
 \hline
Ensemble approach & NA &39.1 &5.09 &7.52 &14.01 \\
\hline
Ensemble approach & GPT2 with trimmed text input &44.55 &7.31 &6.76 &20.59 \\
\hline
Ensemble Approach & GPT2 with trimmed text input + char level language model &42.21 &7.93 &6.72 &21.11 \\
\hline
\end{tabular}
\caption{Adding text completion after every char improves effort saved by 50\%, as compared to text completion only after every word}
\label{table:6}
\end{table}

\begin{table}[H]
\begin{tabular}{|p{2.5cm}||p{1.5cm}|p{1.5cm}|p{1.5cm}|}
\hline
 Model & Data Size (in tokens) &Training Time &P99 Latency in ms \\
 \hline
 pre-trained GPT2 &NA &NA &102  \\
 \hline
 fine-tuned-GPT2 &21.5M &5 Hrs &102 \\
 \hline
 2nd-order Markov Model &21.5M &11m &85 \\
 \hline
 Word level Ensemble (Fine-tuned GPT2 + Markov model) &NA &NA &102 \\
 \hline
 LSTM based char level language model &6.5M &40 mins &62 \\
 \hline
 Word level Ensemble + LSTM based char level LM &NA &NA &80 \\
\hline
\end{tabular}
\caption{Training Time and Latency}
\label{table:7}
\end{table}

From ~\ref{table:7}, it can be seen that fine tuning a GPT2 takes around 5 hours with training data size of 21.5M tokens. This is 1/42 of the time described in \cite{chen2019gmail} . 2nd-order Markov model (ngram model) takes only 11 mins to train and finally a LSTM based character level language model takes around 40 mins to train with training data size of 6.5M tokens. P99 inference latency of overall system is around 80ms in ml.g4dn.xLarge instance in AWS. ml.g4dn.xLarge is the cheapest GPU instance available in aws, which shows the cost-effectiveness of proposed approach.

\subsection{Example suggestions from different models}
\label{sec:robustness}
Table ~\ref{table:8} and ~\ref{table:9} shows some sample output from different models. It can be seen that based on same input text, different model gives different outputs. 

\begin{table}[H]
\begin{tabular}{|p{3cm}|p{2cm}|p{2.5cm}|}
 \hline
 \textbf{Input Text} & \textbf{Model} &\textbf{Suggestion}\\
 \hline
  \hline
   \multirow{5}{*}{I will submit} & LSTM Model & the statements for\\ \cline{2-3}
& Pre-trained GPT2 &a report to the Secretary of State on the results\\ \cline{2-3}
& Fine-tuned GPT2 &the \\ \cline{2-3}
& kth Order Markov Model &start-up cost acct\\ \cline{2-3}
& Ensemble Model  &start-up cost acct\\ \cline{2-3}
  \hline
   \hline
 \multirow{5}{*}{need to adjust} & LSTM Model &the correct account\\ \cline{2-3}
  & Pre-trained GPT2 &the size of the box to fit your needs\\ \cline{2-3}
  & Fine-tuned GPT2 &the chart of accounts as needed\\ \cline{2-3}
  & kth Order Markov Model &the chart of JE into sba portal\\ \cline{2-3}
  & Ensemble Model  &the chart of accounts as needed\\ \cline{2-3}
  \hline
   \hline
\multirow{5}{*}{thank you} & LSTM Model &for the update\\ \cline{2-3}
  & Pre-trained GPT2 &for your support\\ \cline{2-3}
  & Fine-tuned GPT2 &for the update\\ \cline{2-3}
 & kth Order Markov Model &for uploading the requested documents\\ \cline{2-3}
  & Ensemble Model  &for the update\\ \cline{2-3}
  \hline
   \multirow{5}*{\parbox[t]{9.5em}{chase card is a personal card with personal charges, but we start using for payroll (for the points) mid <date>. You will see the paypal charges on that card and that is}} & LSTM Model & not connected to\\ \cline{2-3}
  & Pre-trained GPT2 &the only way to get the card\\ \cline{2-3}
 & Fine-tuned GPT2 &what we used for business \\ \cline{2-3}
 & kth Order Markov Model &paid off in\\ \cline{2-3}
 & Ensemble Model  &what we used for business\\ \cline{2-3}
  \hline
  \hline
\end{tabular}
\caption{Sample suggestion from different models, part-1}
\label{table:8}
\end{table}

\begin{table}[H]
\begin{tabular}{|p{3cm}|p{2cm}|p{2.5cm}|}
   \hline
   \textbf{Input Text} & \textbf{Model} &\textbf{Suggestion}\\
   \hline
  \hline
 \multirow{10}*{\parbox[t]{9.5em}{ all transactions prior to that date will be removed. qbl will remove in a way that we can put them back if needed. smb is sending a spreadsheet indicating}} & LSTM Model & to be paid\\ \cline{2-3}
& Pre-trained GPT2 &the number of transactions that have been sent\\ \cline{2-3}
 & Fine-tuned GPT2 &business expenses paid with personal funds\\ \cline{2-3}
 & kth Order Markov Model &payments since we have permission to move the previous statement\\ \cline{2-3}
& Ensemble Model  &business expenses paid with personal funds\\ \cline{2-3}
 \hline
\end{tabular}
\caption{Sample suggestion from different models, part-2}
\label{table:9}
\end{table}

\begin{itemize}
    \item Pre-trained GPT2 results are generic in nature, while fine-tuned GPT2 results are specific to the domain.
    \item Markov model results are more personalized, when available.
    \item Ensemble model is able to pick up the best suggestion available.
\end{itemize}

\subsection{Production Results}
Since enabling this predictive writing feature for experts, more than a million keystrokes have been saved based on these suggestions. Based on a survey, majority of experts feel that this system helped improve their confidence and efficiency in writing.

\section{Conclusion}
This paper developed a novel ensemble cost-effective approach to auto-complete phrases/sentences/words which can be easily trained and deployed in industry with very less resources. This will improve writing experience and efficiency for Intuit experts in daily life and will save a lot of keystrokes per day. We experimented in detail with different techniques and did extensive evaluation for the choice of our ensemble design. We trained the model by masking user sensitive information to maintain privacy. We also introduced separate word completion along with personalisation and compared performance with other design approaches.

\bibliographystyle{ACM-Reference-Format}
\bibliography{sample-base}

\end{document}